\documentclass[journal,twoside,web]{ieeecolor}
\usepackage{generic}
\usepackage{cite}
\usepackage{amsmath,amssymb,amsfonts}
\usepackage{graphicx}
\usepackage{algorithm,algorithmic}
\usepackage{hyperref}
\hypersetup{hidelinks}
\usepackage{textcomp}
\usepackage{multirow}
\usepackage{booktabs}
\usepackage{array}
\usepackage{bm}

\usepackage[caption=false,font=footnotesize]{subfig}

\newcommand{\mstd}[2]{#1{\scriptsize$\pm$}{\scriptsize #2}}
\newcommand{\bestmstd}[2]{\textbf{#1}{\scriptsize$\pm$}{\scriptsize #2}}
\newcommand{\secondmstd}[2]{\underline{#1}{\scriptsize$\pm$}{\scriptsize #2}}

\def\refname{REFERENCES}
\def\BibTeX{{\rm B\kern-.05em{\sc i\kern-.025em b}\kern-.08em
    T\kern-.1667em\lower.7ex\hbox{E}\kern-.125emX}}
\begin{document}
\title{MPP-GNN: Subject-adaptive Community Detection for fMRI-Based Alzheimer's Disease Classification}
\author{Yang Zhang, Xiao Zhou, Jonathan Warrell, Avram Holmes, Xuan Zhang, and Mark Gerstein
\thanks{Manuscript submitted 29 July 2026. This work was supported in part by the Yale Computational Biology and Bioinformatics M.S. Program Summer Internship Award and in part by research funds from the Gerstein Laboratory, Yale University. (\textit{Yang Zhang, Xiao Zhou, and Jonathan Warrell contributed equally to this work.}) (\textit{Corresponding author: Mark Gerstein.})}
\thanks{Yang Zhang, Xiao Zhou, Jonathan Warrell and Mark Gerstein are with the Program in Computational Biology and Bioinformatics, Yale University, New Haven, CT 06520 USA (e-mail: yang.zhang.yz2483@yale.edu; xiao.zhou@yale.edu; jonathan.warrell@yale.edu; mark@gersteinlab.org).}
\thanks{Avram Holmes is with the Department of Psychiatry, Robert Wood Johnson Medical School, and the Brain Health Institute, Rutgers University, Piscataway, NJ 08854 USA (e-mail: avram.holmes@rutgers.edu).}
\thanks{Xuan Zhang is with School of Design, Pratt Institute, New York, NY 11205 USA (e-mail: xzhang64@pratt.edu).}
}

\maketitle

\begin{abstract}
Functional magnetic resonance imaging (fMRI) is a widely used technique for studying the brain. Recent methods that utilize graph neural networks (GNNs) for analysis of brain functional connectivity have shown great potential for the classification of brain disorders, such as Alzheimer’s
disease (AD). However, these methods often assume a preset number of functional modules across all subjects, which overlooks inter-subject variability. In addition, the discovered modules are rarely used to directly guide the learned connectivity patterns. Here, to address these issues, we propose a Meta Probabilistic Pooling GNN (MPP-GNN). We frame the model’s task as a coupled, bilevel optimization that performs adaptive
graph partitioning hierarchically to discover subject-specific modules and then uses the discovered brain modules as an explicit prior to guide edge refinement and representation learning. We validate MPP-GNN on two public datasets for AD classification, achieving the highest AUC in comparison to established baselines for both datasets. Furthermore, our
analysis demonstrates that MPP-GNN shows significant alignment with the canonical functional-network organization defined by the Yeo brain atlas and reveals a network-level dedifferentiation pattern for AD.
\end{abstract}

\begin{IEEEkeywords}
Brain Network, Graph Neural Network, Deep Learning for Neuroimaging, Alzheimer's Disease, fMRI Biomarker, Graph Structure Learning.
\end{IEEEkeywords}

\section{Introduction}
\label{sec:introduction}
\IEEEPARstart{A}{lzheimer's} disease (AD) is a prevalent neurodegenerative disorder characterized by progressive cognitive decline. Amyloid-$\beta$ plaques and tau neurofibrillary tangles are two neuropathological hallmarks that accumulate in the brain years to decades before the onset of clinical symptoms. This long preclinical window motivates early detection and intervention before irreversible neuronal loss occurs. Among the available detection approaches, functional magnetic resonance imaging (fMRI) provides a non-invasive technique that can capture disruptions in inter-regional neural communication and is widely available in clinical settings~\cite{ADReview}.

Neuroimaging data from fMRI can be used to model the brain network as a graph. In the standard pipeline, a brain atlas partitions preprocessed fMRI volumes into regions of interest (ROIs) that can be viewed as nodes. The pairwise statistical dependencies between the ROI-level blood-oxygen-level-dependent (BOLD) time series are then estimated to form a functional connectivity (FC) matrix that defines weighted edges~\cite{BrainGB}. Among various models, graph neural networks (GNNs) are well suited to this representation, and have received increasing attention recently. Compared with traditional machine learning models, GNNs can explicitly utilize both node attributes and relational structure through the message passing mechanism. The use of GNNs to model brain graphs has not only demonstrated state-of-the-art performance in various brain disorder classification tasks, but also serves as a crucial analytical tool for identifying biomarkers associated with neurological disorders~\cite{brain-graph-survey}. 

Despite these advances, GNN-based brain graph analysis faces several challenges: (1) BOLD-derived FC matrices are inherently noisy~\cite{noise-contribution}. Non-neural noise introduced by physiological artifacts and limited scan duration cause large variations within a single subject and across different subjects. (2) The human brain is highly modular. Studies have shown that  AD is associated with disrupted modular organization, with substantial variation across individuals and disease stages~\cite{ADReview, ad-meta-review}. However, many GNN-based community detection methods impose a preset clustering capacity shared across subjects, which conflicts with the heterogeneous nature of brain organizations~\cite{BrainNetTransformer, BrainGNN, transformer-hierarchical}. (3) Interpretability in brain graph models remains challenging. Common post-hoc methods, such as saliency maps and surrogate explanations, often lack consistency across different explanation algorithms and offer limited transparency into the model's internal mechanisms. 

In fact, these challenges are closely related. Community structure can guide edge refinement by distinguishing intra-community from inter-community connections. Refined edges and learned representations can in turn improve the partition strategy. Jointly addressing them within a single framework, rather than treating them as independent preprocessing or post-hoc steps, can lead to more robust and interpretable brain graph models.

To address these challenges, we propose the Meta Probabilistic Pooling Graph Neural Network (MPP-GNN), an end-to-end framework that jointly refines noisy FC matrices, adaptively discovers subject-specific community structure, and produces inherent interpretability. Our main contributions are summarized as follows.
\begin{itemize}

\item We formulate brain graph classification as a bilevel optimization problem, where the outer level learns an adaptive community detection strategy to guide the inner-level joint optimization of edge refinement and node representation learning. This provides a principled mechanism to couple community detection and edge denoising with the final classification task within a single end-to-end framework.

\item We design an Affinity-Propagation-based Hierarchical Pooling Module (AP-HPM) that discovers multi-resolution community structure without a preset number of clusters. We also introduce a Probabilistic Edge Refinement Module (PERM), which uses the discovered community structure as a subject-specific structural prior to estimate edge-retention probabilities. Experiments on two public datasets for AD classification demonstrate that MPP-GNN achieves superior performance compared to multiple established baselines across different metrics.

\item We show that AP-HPM can recover canonical functional network organization consistent with the Yeo brain atlas. Similarly, PERM can produce consistent edge-retention patterns. Both modules reveal AD-related network dedifferentiation, aligning with established neuroscience literature.

\end{itemize}
 
\section{Related Work}

\subsection{Graph Structure Learning}

Graph Structure Learning (GSL) jointly optimizes graph topology and node representations, and has been widely applied to denoise noisy relational data~\cite{graph-structual-learning-survey}. Existing GSL methods can be broadly categorized by how they model edge weights. Metric-based approaches derive edge weights from pairwise node similarities, using trainable kernels such as Mahalanobis distance~\cite{AGCN}, cosine similarity with learnable parameters~\cite{IDGL,GNN-Guard}, or inner products~\cite{GRCN, GAUG-M, CAGNN}. Neural approaches directly apply neural networks to model edge weights. For example, GLCN~\cite{GLCN} uses a single-layer neural network, and NeuralSparse~\cite{NeuralSparse} leverages multilayer perceptrons to learn edge connectivity strength. Many methods also leverage the attention mechanism. GAT~\cite{GAT} first introduces the attention mechanism to GNNs through the use of masked self-attention over one-hop neighborhoods. Transformer-like full-attention architectures have also been generalized to the graph domain~\cite{Graphormer}.
Direct approaches treat the adjacency matrix itself as a learnable variable, often with regularization terms that encourage sparsity~\cite{sparsity-regularizer} and smoothness~\cite{smoothness-regularizer}. In the context of brain graph learning, the FC matrix encodes the connectivity pattern among different regions of the brain that we aim to recover from the preprocessing noise and artifacts. However, conventional graph structure learning methods do not explicitly incorporate community structure as an inductive bias for edge refinement. In addition, attention weights alone are not guaranteed to provide faithful
explanations of individual predictions~\cite{attention-is-not-explanation}.

\subsection{Graph Pooling}

Graph pooling is a method used in graph neural networks to reduce the complexity of graph data~\cite{graph-pooling-survel}. Early pooling methods rely on graph-partition objectives. Spectral clustering uses eigendecomposition to relax this objective to continuous eigenvectors of the graph Laplacian. Graclus~\cite{graclus} directly approximates the same objective through multilevel coarsening and greedy matching. Global pooling aggregates all node features into a single vector via summation, averaging, or maximization. Attention mechanisms have also been used to enhance pooling operations~\cite{attention-on-pooling}. However, global pooling operations have been criticized for ignoring the rich hierarchical structure information inside the graph. Hierarchical pooling progressively constructs coarsened graphs across layers. DiffPool~\cite{diffpool} learns a dense soft assignment matrix from a one-layer GNN to assign nodes to clusters. EigenPool~\cite{EigenPool} uses the graph Fourier transform to aggregate nodes during the hierarchical pooling. StructPool~\cite{StructPool} designs conditional random fields to incorporate higher-order information among different nodes. Other methods rely on a scoring function to keep only a certain number of important nodes during hierarchical pooling to ensure efficiency. SAGPool~\cite{SAGPool} uses the self-attention to score and retain nodes. However, many hierarchical pooling methods impose a preset pooling capacity, such as a fixed number of clusters or retained nodes, which may limit adaptation to subject-specific organization. Adaptive community detection algorithms such as the Louvain method avoid this constraint but are typically applied as a fixed preprocessing step that cannot be refined jointly with the downstream learning objective.

\subsection{fMRI Functional Brain Networks}

Recent fMRI-specific graph models have increasingly recognized the importance of modular brain organization for disease prediction and interpretation~\cite{brainGNN-survey, modular-brain, community-is-important}. However, existing methods remain limited in how the community structure is modeled. Some approaches impose a fixed clustering capacity. BrainNetTF~\cite{BrainNetTransformer} introduces an orthonormal cluster readout, but the number of clusters is selected as a predefined hyperparameter. THC~\cite{transformer-hierarchical} further extends this idea to hierarchical clustering, but its layer-wise cluster sizes are also preset. Community-aware transformer variants such as Com-BrainTF~\cite{Com-BrainTF} and CAGT~\cite{CAGT} likewise rely on predefined or externally constructed community structure to guide representation learning. Some methods perform community extraction as a separate preprocessing step. For example, SW-HGL~\cite{SW-HGL} applies Louvain-based clustering to obtain micro-scale communities prior to the predictive model. However, this prevents community detection from being refined by downstream supervision. Meanwhile, graph structure learning and interpretable models such as MVS-GCN~\cite{mvs-gcn} and IBGNN~\cite{IBGNN} can suppress noisy connections, but do not use community structure to guide edge refinement. These limitations motivate a framework that jointly learns adaptive community partitions, community-guided edge denoising, and hierarchical graph representations within a single end-to-end pipeline.

\begin{figure*}[!t]
\centering
\includegraphics[width=0.98\textwidth]{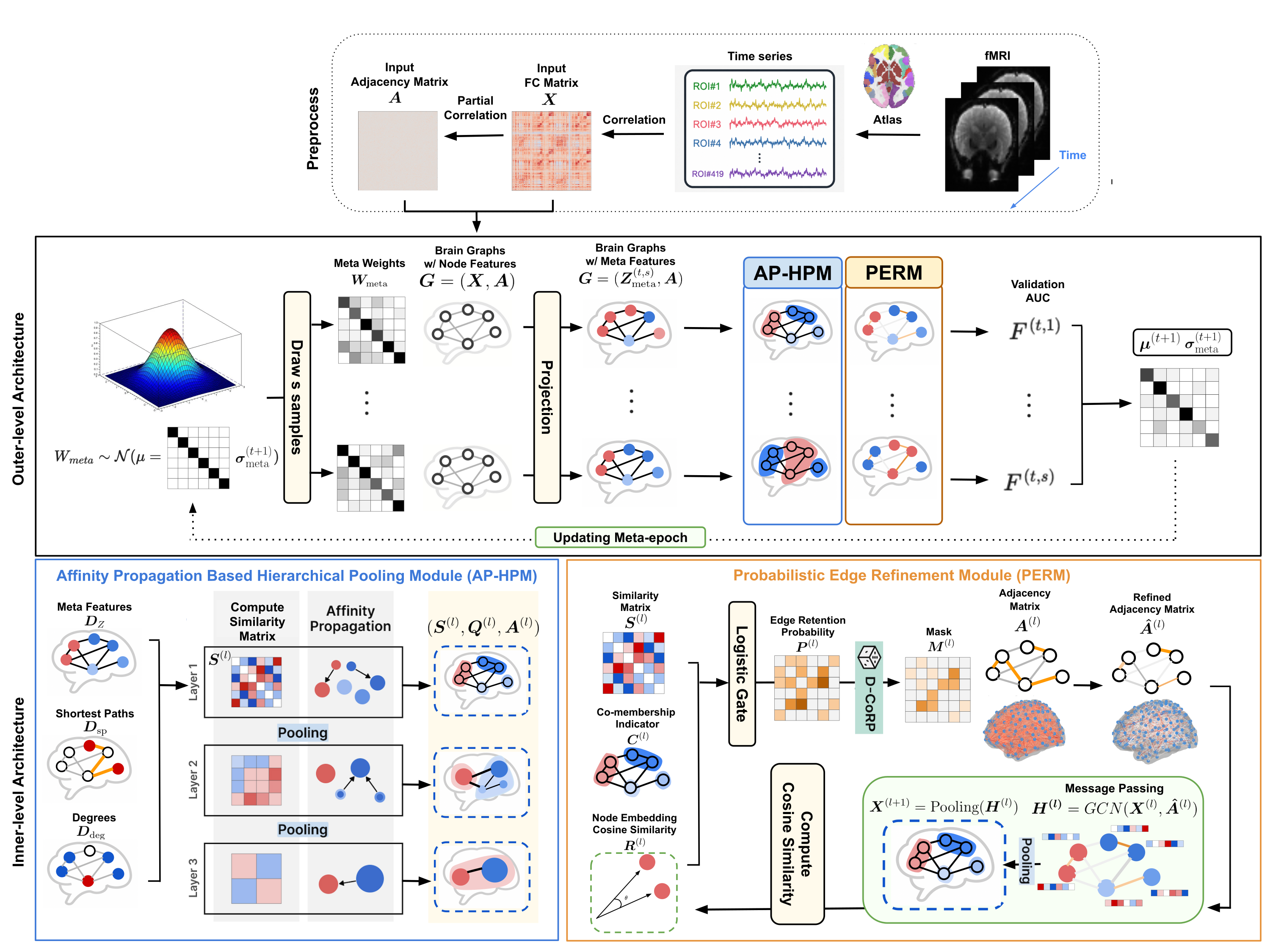}
\caption{The architecture of MPP-GNN, where the top row illustrates the data preprocessing, the middle row depicts the outer-level optimization and the bottom row details the inner-level AP-HPM and PERM modules.}
\label{fig:architecture}
\end{figure*}

\section{Methodology}

\subsection{Preliminaries}

Define an undirected, weighted graph as $G = (\bm{X}, \bm{A})$, where $\bm{X} \in \mathbb{R}^{n \times d}$ is the node feature matrix with $n$ nodes and $d$ dimensions, and $\bm{A} \in \mathbb{R}^{n \times n}$ is the weighted adjacency matrix. Given a graph dataset $\mathcal{D}=\{(G_1, y_1), (G_2,y_2),\dots\}$ where $y_i\in \mathcal{Y}$ is the disease label corresponding to graph $G_i\in \mathcal{G}$, our goal of graph classification is to learn a model $f_{(\theta, \phi)}:\mathcal{G}\rightarrow\mathcal{Y}$ that maps each graph to a label. At a high level, AP-HPM first pre-computes the hierarchical graph structures for all layers before the inner-level optimization. Then, PERM refines the adjacency matrix at each layer, a graph convolutional network (GCN) updates the node representations, and AP-HPM pools the graphs according to the pre-computed community assignments.

\noindent\textbf{Bilevel Optimization.} We frame brain graph classification as a bilevel optimization, where the outer level learns an optimal graph-partitioning strategy from Affinity Propagation~\cite{affinity-propagation}, and the inner level optimizes hierarchical representations based on the clustering results. The central challenge is that graph partitioning produces discrete cluster
assignments that break the gradient flow to downstream inner-level GNN parameters. A common approach is to fix the graph partition as a preprocessing step as in~\cite{SW-HGL}, but this can
decouple community discovery from representation learning. We instead adopt the smoothing-based variational optimization (SBO) strategy~\cite{smoothing-based-optimizer, gpgnn} that bridges the community discovery and representation learning through distributional parameterization.

Denote $\theta$ as the learnable inner-level parameters and $\phi := (\bm{\mu}, \sigma_{\mathrm{meta}})$ as the outer-level parameters that parameterize a multivariate Gaussian distribution over the partition-inducing meta weights, 
\begin{equation}
    \bm{W}_{\mathrm{meta}} \sim \mathcal{N}(\bm{\mu},\sigma_{\mathrm{meta}}^2 \bm{I}).
\end{equation}
For the $s$-th sampled meta weight $\bm{W}_{\mathrm{meta}}^{(t,s)}$ at meta-epoch $t$, the input node features $\bm{X}^{(0)}$ are projected into a meta-feature space, 
\begin{equation}
    \bm{Z}_{\mathrm{meta}}^{(t,s)} = \bm{X}^{(0)} \bm{W}_{\mathrm{meta}}^{(t,s)}.
\end{equation}
The resulting meta-features are used to construct the first-layer similarity matrix $\bm{S}^{(t,s,1)}$ for AP-HPM, which is then used as input to affinity propagation. Thus, $\bm{W}_{\mathrm{meta}}^{(t,s)}$ determines the partition strategy used during the corresponding inner-level optimization. 

Our main objective is to learn $\phi$ by iteratively refining the distribution $\mathcal{N}(\bm{\mu},\sigma_{\mathrm{meta}}^2 \bm{I})$ toward partition strategies that maximize model performance. 
Formally,

\begin{equation}
 \phi^* =
 \arg\max_{\phi}\,
 \mathbb{E}_{\bm{W}_{\mathrm{meta}}
   \sim \mathcal{N}(\bm{\mu},\sigma_{\mathrm{meta}}^2 \bm{I})}
 \big[\,F(\theta^*(\bm{W}_{\mathrm{meta}}))\,\big],
\end{equation}
\begin{equation}
    \quad\text{s.t.}\quad
\theta^*(\bm{W}_{\mathrm{meta}})
=
\arg\min_{\theta}
\mathcal{L}(\theta;\, \bm{W}_{\mathrm{meta}}).
\end{equation}
Here, $\mathcal{L}$ is the inner-level loss, and $F$ is the scalar validation score evaluated at the optimized inner-level parameters $\theta^*(\bm{W}_{\mathrm{meta}})$. For a sampled 
$\bm{W}_{\mathrm{meta}}$, the inner level learns the model parameters under the corresponding partition structure. The outer level then updates $\phi$ so that the distribution over $\bm{W}_{\mathrm{meta}}$ assigns higher probability to partition strategies that produce better validation performance. 

At the algorithmic level, at each meta-epoch $t$, we draw $S$ samples of meta-weights from the multivariate Gaussian distribution. Each sample induces a distinct AP partition strategy that imposes structural constraints on inner-level learning. The inner-level then trains to convergence, producing optimal parameters $\theta^{*(t,s)}$ and a corresponding score $F^{(t,s)}$. The outer-level parameters are updated as follows:

\begin{equation}
    \bm{\mu}^{(t+1)}
    = \frac{\sum_s F^{(t,\,s)}\,\bm{W}_{\mathrm{meta}}^{(t,\,s)}}
           {\sum_s F^{(t,\,s)}},
    \label{mu_update}
\end{equation}

\begin{equation}
    \sigma_{\mathrm{meta}}^{(t+1)}
    =\sqrt{
      \frac{\sum_s F^{(t,\,s)}\,
            \|\bm{W}_{\mathrm{meta}}^{(t,\,s)}-\bm{\mu}^{(t+1)}\|_2^2}
           {N_{\bm{W}_{\mathrm{meta}}}\,\sum_s F^{(t,\,s)}}
    }.
    \label{sigma_update}
\end{equation}
where $N_{\bm{W}_{\mathrm{meta}}}$ is the dimensionality of $\bm{W}_{\rm{meta}}$. The
outer-level optimization ends when $\sigma_{\mathrm{meta}}$ falls below a convergence threshold $\epsilon$ or after $T$ meta-epochs.

SBO treats the outer objective as a black-box function and only requires pointwise evaluation of a non-negative score, without assuming differentiability~\cite{smoothing-based-optimizer}. The area under the receiver operating characteristic curve (AUC) metric is particularly appropriate for the class-imbalanced clinical neuroimaging setting and is naturally bounded in $[0,1]$. It can be directly used in the score-weighted updates of Eq.~\eqref{mu_update} and Eq.~\eqref{sigma_update}. This choice is also consistent with the standard bilevel optimization paradigm, where the inner level minimizes the training objective while the outer level selects structure-related variables according to the validation performance~\cite{DARTS}.

\noindent\textbf{Graph Neural Network.} We adopt Graph Convolutional Networks (GCNs)~\cite{GCN} as the message propagation operation for our model. For brevity, we omit the meta-optimization indices $(t, s)$ in the following discussions. Unless otherwise stated, all subsequent derivations are presented for a single sample within one meta-epoch. The node embedding at layer $l$, denoted as $\bm{H}^{(l)}$, is computed through a sequence of graph convolution, normalization, and non-linear activation:

\begin{equation}
\bm{H}^{(l)} = \mathrm{GELU}\!\Big(\mathrm{norm}\!\big(
\big(\hat{\bm{D}}^{(l)}\big)^{-\frac{1}{2}}
\hat{\bm{A}}^{(l)}
\big(\hat{\bm{D}}^{(l)}\big)^{-\frac{1}{2}}
  \bm{X}^{(l)} \bm{W}^{(l)} + \bm{b}^{(l)}
\big)\Big).
\label{mp}
\end{equation}
where $\hat{\bm{A}}^{(l)}=\mathrm{PERM}(\tilde{\bm{A}}^{(l)})$ is the refined adjacency matrix, $\tilde{\bm{A}}^{(l)}=\bm{A}^{(l)}+\bm{I}$ denotes the adjacency matrix with self-loops as in the standard GCN, $\hat{D}_{ii}^{(l)} = \sum_j \hat{A}_{ij}^{(l)}$, and $\bm{W}^{(l)}$ is a trainable weight matrix. $\bm{X}^{(l+1)}=\mathrm{AP\text{-}HPM}(\bm{H}^{(l)})$ is the pooled node features before the message passing, serving as the input node features for the message passing at layer $l+1$. $\mathrm{norm}$ denotes $\mathrm{GraphNorm}(\cdot)$, which applies instance-level normalization centered on the graph~\cite{GraphNorm}. $\mathrm{GELU}(\cdot)$ is the Gaussian Error Linear Unit activation function~\cite{GELU}.

\subsection{Affinity Propagation based Hierarchical Pooling}

\noindent\textbf{Similarity Matrix Projection.} At the first layer, we construct the initial similarity matrix $\bm{S}^{(1)}$ by combining learnable meta-features $\bm{Z}_{\mathrm{meta}}^{(t,s)}$ with fixed graph topological information. We define: 

\begin{equation}
\bm{S}^{(1)} \;=\; \exp\!\big(
    -\bm{D}_Z
    \;-\; \lambda_{\mathrm{deg}}\,\bm{D}_{\mathrm{deg}}
    \;-\; \lambda_{\mathrm{sp}}\,\bm{D}_{\mathrm{sp}}
\big).
\end{equation}
where $\bm{D}_Z$ is the pairwise $\ell_1$ distance in the meta-feature space, $\bm{D}_{\mathrm{deg}}$ is the absolute degree difference matrix, and $\bm{D}_{\mathrm{sp}}$ is the shortest-path distance matrix. The topological terms $\bm{D}_{\mathrm{deg}}$ and $\bm{D}_{\mathrm{sp}}$ are computed once
from the input graph and remain fixed. These two terms encode how similarly connected two ROIs are and how far apart they lie in the graph topology, respectively. The meta-feature term $\bm{D}_Z$, on the other hand, evolves across meta-epochs as the outer-level optimization refines $\bm{W}_{\mathrm{meta}}$. $\lambda_{\mathrm{deg}}$ and $\lambda_{\mathrm{sp}}$ are two
hyperparameters that determine the importance of each topological feature.

\noindent\textbf{Hierarchical Pooling of Graphs.} Given the similarity matrix $\bm{S}^{(l)}$, we apply the Affinity Propagation (AP) algorithm to partition the $n^{(l)}$ nodes into clusters. AP accepts arbitrary similarity matrices, which allows the topological and meta-feature terms in $\bm{S}^{(l)}$ to be flexibly combined. Because AP does not require a predetermined cluster number, $n^{(l+1)}$ adapts to the structural complexity of each individual graph. We denote $\bm{Q}^{(l)} \in \{0,1\}^{n^{(l)} \times n^{(l+1)}}$ as the resulting binary assignment matrix that assigns each node to its cluster at layer $l$. The AP preference is treated as a hyperparameter and tuned.

The AP-HPM module coarsens the graph by aggregating node features and inter-cluster connections. Let $\bm{D}_{q}^{(l)} = \mathrm{diag}(\bm{Q}^{(l)\top} \mathbf{1})$ denote the diagonal matrix of the cluster sizes. We compute the pooled node features
by averaging within each cluster:
\begin{equation}
\bm{X}^{(l+1)} \;=\;
  \big(\bm{D}_{q}^{(l)}\big)^{-1}\,\bm{Q}^{(l)\top}\,\bm{H}^{(l)}
  \;\in\; \mathbb{R}^{n^{(l+1)} \times d^{(l)}},
\end{equation}
where $\bm{H}^{(l)}$ is the node embedding after the GCN message passing and $d^{(l)}$ denotes the feature dimension at layer $l$. For the pooled adjacency, AP-HPM constructs a sparse inter-cluster graph. Let $\mathcal{E}^{(l)}_{k_1,k_2}$ denote the set of edges in $G^{(l)}$ that connect two distinct clusters $k_1$ and $k_2$ ($k_1 \neq k_2$). The pooled edge weight between two distinct clusters is computed as the
mean weight of the observed inter-cluster edges:
\begin{equation}
A^{(l+1)}_{k_1,k_2}
=
\frac{1}{|\mathcal{E}^{(l)}_{k_1,k_2}|}
\sum_{(i,j)\in \mathcal{E}^{(l)}_{k_1,k_2}} A^{(l)}_{ij},
\quad k_1 \neq k_2,
\end{equation}
with $A^{(l+1)}_{k_2,k_1}=A^{(l+1)}_{k_1,k_2}$ for undirected graphs. The intra-cluster edges are not explicitly preserved as diagonal entries in $\bm{A}^{(l+1)}$. Instead, they are 
introduced through the GCN self-loop term in 
$\tilde{\bm{A}}^{(l)}$, whose effective strength is modeled by the layer-wise PERM gate during inner-level training. If $|\mathcal{E}^{(l)}_{k_1,k_2}|=0$, the  corresponding pooled edge is assigned zero weight and is omitted from the sparse coarsened graph. For layers $l > 1$, AP-HPM also pools the similarity matrix hierarchically by averaging over all node pairs within each cluster pair:
\begin{equation}
\bm{S}^{(l+1)}
\;=\;
\big(\bm{D}_{q}^{(l)}\big)^{-1}\,\bm{Q}^{(l)\top}\,
\bm{S}^{(l)}\,
\bm{Q}^{(l)}\,\big(\bm{D}_{q}^{(l)}\big)^{-1}.
\end{equation}
This cascade produces a multi-resolution hierarchy from ROIs to networks to systems.

Importantly, structural components ($\bm{S}^{(l)}, \bm{Q}^{(l)}, \bm{A}^{(l)}$) are derived from sampled meta-weights $\bm{W}_{\mathrm{meta}}$ and remain static throughout each inner-level training phase. Thus, they can be pre-computed at the onset of each inner-level optimization and remain fixed throughout that training phase to ensure efficiency. In contrast, the pooled node features $\bm{X}^{(l)}$ evolve dynamically with the inner-level GCN parameters $\theta$. The refined adjacency $\hat{\bm{A}}^{(l)}$ also evolves. Its connectivity pattern is inherited from the pre-computed $\bm{A}^{(l)}$, but its edge weights are continuously re-estimated by PERM during inner-level training. As the outer-loop updates $\phi$ through meta-epochs, the distribution of $\bm{W}_{\mathrm{meta}}^{(t,s)}$ shifts, generating progressively refined partition strategies that guide inner-level learning.

\subsection{Probabilistic Edge Refinement}

FC matrices contain false correlations arising from physiological noise and preprocessing artifacts. Such noise propagates through message passing as shown in \eqref{mp} and pollutes potential signals. Traditional deterministic thresholding discards weak but potentially informative edges, while learning soft attention masks raises concerns about the trustworthiness of the resulting interpretations. PERM addresses these conflicts by learning a layer-wise logistic gate whose inputs have explicit structural meanings. At each hierarchical level $l$, the gate combines the structural priors obtained from AP-HPM with inner-level node representations to produce an interpretable score matrix $\bm{P}^{(l)}$, where each entry $P_{ij}^{(l)}\in(0,1)$ controls the strength with which edge $(i,j)$ participates in message passing. Each gate coefficient directly reflects the contribution of a specific factor to edge retention. A differentiable stochastic mask is then sampled from the learned probability matrix to sparsify the edges.

\noindent\textbf{Global Logistic Edge Gate.} Once we obtain the structural components ($\bm{S}^{(l)}, \bm{Q}^{(l)}, \bm{A}^{(l)}$) of each layer from AP-HPM, we construct three edge-level features for every pair $(i,j)$ of nodes: (i)~the similarity metric $S^{(l)}_{ij}$ encoding meta-feature proximity and
topological affinity; (ii)~a co-membership indicator $C^{(l)}_{ij}\in \{-1, +1\}$ derived from the cluster assignment $\bm{Q}^{(l)}$, where $C^{(l)}_{ij}=+1$ if nodes $i$ and $j$ belong to the same cluster and $-1$ otherwise; and (iii)~the cosine similarity $\bm{R}^{(l)}$ between their node features $\bm{x}_i^{(l)}$ and $\bm{x}_j^{(l)}$ in the current layer. We use a layer-wise logistic model to map these features to an edge-retention probability:
\begin{equation}
\bm{P}^{(l)} = \sigma\!\big(
  \lambda_1^{(l)}\,\bm{S}^{(l)}
  +\lambda_2^{(l)}\,\bm{C}^{(l)}
  +\lambda_3^{(l)}\, \bm{R}^{(l)}
  +\lambda_4^{(l)}
\big).
\end{equation}
where $\sigma$ denotes the sigmoid function and $\lambda_1^{(l)}$ to $\lambda_4^{(l)}$ are learnable scalar parameters. Because $\bm{S}^{(l)}$ and $\bm{C}^{(l)}$ are fixed within each inner loop, PERM inherits the community structure discovered by AP-HPM. Intra-cluster edges receive a positive bias through $\bm{C}^{(l)}$, while inter-cluster edges must compensate through high feature-level or meta-feature similarity to remain influential. Meanwhile, the pairwise cosine similarity of node features $\bm{R}^{(l)}$ evolves dynamically as the GCN updates its representations, which allows edge retention to adapt to the learned features. In other words, an edge can still receive a high retention probability $P^{(l)}_{ij}$ when the two incident nodes have similar learned representations, even if their fixed structural prior is weak. The scalar parameterization keeps the gate highly interpretable. Each coefficient directly quantifies the relative contribution of community membership, meta-feature similarity, and
feature-level similarity to edge retention.

\noindent\textbf{Stochastic Bernoulli Mask.} To convert the continuous probabilities $\bm{P}^{(l)}$ into a near-discrete mask while preserving gradient flow, we adapt a relaxed Bernoulli sampling scheme~\cite{D-CoRP}. At each forward pass, a random matrix $\bm{U}^{(l)}$ is drawn element-wise from $\mathcal{U}(0,1)$, and the mask is computed as:
\begin{equation}
\bm{M}^{(l)}=\frac{1}{2}\left(
  \tanh\!\left(\frac{\bm{P}^{(l)}-\bm{U}^{(l)}}{\tau}\right)+1
\right).
\end{equation}
where $\tau$ is a temperature hyperparameter. As $\tau \rightarrow 0$, $\bm{M}^{(l)}$ converges to a discrete $\{0, 1\}$ Bernoulli mask. For larger $\tau$, it acts as a soft, continuous gate that attenuates rather than removes edges. Edges with higher probability $P_{ij}^{(l)}$ are retained with greater likelihood, effectively imposing a soft sparsity constraint without a hard threshold. Stochastic sampling also acts as structural data augmentation
during training. Every forward pass operates on a slightly different subgraph, which regularizes the model and mitigates overfitting.

The refined adjacency matrix is then obtained via element-wise
multiplication:
\begin{equation}
\hat{\bm{A}}^{(l)} = \tilde{\bm{A}}^{(l)} \odot \bm{M}^{(l)},
\end{equation}
on which message passing \eqref{mp} is performed to produce the node
embeddings $\bm{H}^{(l)}$.

\subsection{Loss Function of Inner-level Base Model}

In addition to the classification loss $\mathcal{L}_{\mathrm{cls}}$ for
graph classification, we introduce a regularization term
$\mathcal{L}_{\mathrm{edge}}$ on the edge-retention probability matrices produced by PERM. $\mathcal{L}_{\mathrm{edge}}$ encourages edge-retention probabilities $P^{(l)}_{ij}$ to move away from uncertain values around $0.5$ and commit to near-binary decisions. We define the binary entropy of a single probability and average it over all candidate
edges and hierarchical levels:
\begin{equation}
\mathcal{L}_{\mathrm{edge}}
=
\frac{1}{L}
\sum_{l=1}^{L}
\frac{1}{|\mathcal{E}^{(l)}|}
\sum_{(i,j)\in \mathcal{E}^{(l)}}
\mathrm{entropy}\!\left(P_{ij}^{(l)}\right).
\end{equation}
where $L$ is the number of hierarchical levels, $P_{ij}^{(l)}\in(0,1)$ is
the edge-retention probability for edge $(i,j)$ at layer $l$, and
$\mathcal{E}^{(l)}$ denotes the set of candidate edges in $\tilde{\bm{A}}^{(l)}$ on which PERM operates at layer $l$. Minimizing $\mathcal{L}_{\mathrm{edge}}$ encourages near-binary edge decisions, hence improving the interpretability and sharpness of the learned graph refinement.

Our final training objective can be written as:
\begin{equation}
\mathcal{L}
=
\mathcal{L}_{\mathrm{cls}}
+
\lambda_{\mathrm{edge}}\,\mathcal{L}_{\mathrm{edge}},
\end{equation}
where $\lambda_{\mathrm{edge}}$ is a hyperparameter that controls the strength of the edge entropy
regularization. $\mathcal{L}_{\mathrm{cls}}$ is a classification loss
adapted to the label structure of each dataset.

\begin{table}[t]
\caption{Class distribution of the datasets.}
\label{tab:dataset_distribution}
\centering
\footnotesize
\setlength{\tabcolsep}{6pt}
\setlength{\aboverulesep}{0.15ex}
\setlength{\belowrulesep}{0.15ex}
\renewcommand{\arraystretch}{0.94}
\begin{tabular}{
    >{\centering\arraybackslash}p{1.65cm}
    >{\centering\arraybackslash}p{1.25cm}
    >{\centering\arraybackslash}p{1.65cm}
}
\toprule
Dataset & Class & \# Subjects \\
\midrule
\multirow{2}{*}{UK Biobank}
& CN & 246 \\
& AD & 54 \\
\midrule
\multirow{3}{*}{ADNI}
& CN  & 365 \\
& MCI & 247 \\
& AD  & 68 \\
\bottomrule
\end{tabular}
\end{table}

\section{Experimental Studies}

\subsection{Dataset and Preprocessing}
We evaluated the MPP-GNN on two public datasets, UK Biobank~\cite{ukbiobank} and ADNI (Alzheimer's Disease Neuroimaging Initiative)~\cite{ADNI3}. For both datasets, we used the Schaefer 2018 atlas~\cite{schaefer2018} to define 400 cortical regions and included 19 additional subcortical regions following the preprocessing pipeline in~\cite{ukbiobank_pipeline}, yielding 419 ROIs in total. We applied PCA to reduce node features $X^{(0)}$ to 32 before feeding them into the inner-level base model. For edge weights, we calculated partial correlations and kept the top $30\%$ of correlations.

\subsubsection{UK Biobank Dataset} UK Biobank is a population epidemiology study of 500,000 participants aged 40 to 69 years, recruited between 2006 and 2010. A subset of 100,000 participants is being recruited for multimodal imaging, including brain structural MRI and resting-state fMRI (rs-fMRI). Here, we used the preprocessed 37,848 FC matrices of different participants, and the detailed pipeline is described in~\cite{ukbiobank_pipeline}. In data field 42020, 5,209 participants were identified as Alzheimer's disease cases. Among them, only 54 had brain rs-fMRI scans available. Given the substantial class imbalance in the full cohort and the computational burden of large-scale graph learning, we retained all 54 AD subjects and randomly sampled 246 cognitively normal (CN) controls for downstream experiments. The final dataset therefore contained 300 subjects in total.

\subsubsection{ADNI Dataset}
ADNI is a longitudinal, multicenter observational study that provides imaging, clinical, genetic, and biomarker data for Alzheimer's disease research~\cite{ADNI3}. In this study, we used participants from ADNI Phase 3. We included three diagnostic groups: cognitively normal (CN), mild cognitive impairment (MCI), and Alzheimer's disease (AD). We utilized the fMRIPrep~\cite{fMRIPrep} pipeline to preprocess the raw structural and functional MRI data into BOLD images, which includes the step of skull-stripping, tissue segmentation, motion correction, EPI-to-T1w registration, spatial normalization, and confound estimation.

\begin{table*}[t]
\caption{Classification results over five-fold cross-validation (mean $\pm$ standard deviation, \%). The best and second-best results are shown in \textbf{bold} and \underline{underlined}, respectively.}
\label{tab:main_results}
\centering
\footnotesize
\setlength{\tabcolsep}{2.4pt}
\setlength{\aboverulesep}{0.15ex}
\setlength{\belowrulesep}{0.15ex}
\renewcommand{\arraystretch}{0.93}
\begin{tabular}{
    >{\centering\arraybackslash}p{1.65cm}
    >{\raggedright\arraybackslash}p{2.45cm}
    >{\centering\arraybackslash}p{1.42cm}
    >{\centering\arraybackslash}p{1.42cm}
    >{\centering\arraybackslash}p{1.42cm}
    >{\centering\arraybackslash}p{1.42cm}
    >{\centering\arraybackslash}p{1.42cm}
    >{\centering\arraybackslash}p{1.42cm}}
\toprule
& & \multicolumn{3}{c}{UK Biobank} & \multicolumn{3}{c}{ADNI} \\
\cmidrule(lr){3-5}\cmidrule(lr){6-8}
Category & Model & AUC & ACC & F1 & AUC & ACC & F1 \\
\midrule
\multirow{2}{*}{\parbox[c]{1.65cm}{\centering\itshape Conventional\\ ML}}
& SVM           & \secondmstd{74.13}{10.34} & \mstd{67.67}{8.47} & \mstd{70.39}{6.54} & \secondmstd{70.04}{3.36} & \secondmstd{58.38}{1.10} & \secondmstd{54.31}{1.04} \\
& Random Forest & \mstd{65.00}{9.17} & \mstd{61.00}{14.97} & \mstd{64.07}{13.43} & \mstd{68.69}{2.19} & \mstd{55.29}{1.57} & \mstd{44.65}{2.25} \\
\midrule
\multirow{7}{*}{\parbox[c]{1.65cm}{\centering\itshape General-\\purpose GNNs}}
& GCN       & \mstd{52.87}{5.57}  & \mstd{51.33}{11.99} & \mstd{55.35}{13.48} & \mstd{68.07}{4.45} & \mstd{51.03}{3.96} & \mstd{49.59}{4.49} \\
& GAT       & \mstd{55.58}{8.15}  & \mstd{58.00}{9.63}  & \mstd{60.94}{7.04}  & \mstd{68.06}{4.27} & \mstd{54.85}{4.55} & \mstd{53.98}{5.62} \\
& GIN       & \mstd{56.13}{7.08}  & \mstd{60.33}{9.33}  & \mstd{62.78}{7.34}  & \mstd{66.71}{5.65} & \mstd{52.94}{5.83} & \mstd{53.88}{5.15} \\
& GraphSAGE & \mstd{55.02}{10.38} & \mstd{67.33}{4.03}  & \mstd{69.41}{3.53}  & \mstd{65.88}{2.52} & \mstd{52.65}{4.66} & \mstd{48.06}{4.95} \\
& DiffPool  & \mstd{58.77}{10.70} & \mstd{63.67}{10.19} & \mstd{66.06}{7.70}  & \mstd{63.79}{5.02} & \mstd{46.47}{6.33} & \mstd{48.09}{5.60} \\
& TopKPool  & \mstd{48.03}{9.99}  & \mstd{54.67}{16.24} & \mstd{56.38}{15.47} & \mstd{62.05}{2.53} & \mstd{48.24}{2.89} & \mstd{49.05}{2.49} \\
& SAGPool   & \mstd{44.26}{9.62}  & \mstd{57.67}{10.83} & \mstd{60.89}{8.65}  & \mstd{55.52}{4.13} & \mstd{37.94}{10.12} & \mstd{39.79}{9.24} \\
\midrule
\multirow{6}{*}{\parbox[c]{1.65cm}{\centering\itshape Brain-network\\ models}}
& BrainNetTF    & \mstd{68.70}{10.84} & \mstd{72.00}{9.03} & \mstd{71.02}{4.76} & \mstd{65.84}{2.40} & \mstd{42.79}{8.03} & \mstd{30.22}{13.09} \\
& BrainNetCNN   & \mstd{62.77}{9.84}  & \mstd{63.00}{9.45} & \mstd{66.13}{7.67} & \mstd{69.99}{5.16} & \mstd{56.62}{5.03} & \mstd{53.68}{5.05} \\
& BrainGNN      & \mstd{45.18}{4.87}  & \mstd{46.33}{17.68} & \mstd{48.47}{13.71} & \mstd{49.59}{5.46} & \mstd{36.32}{0.36} & \mstd{19.36}{0.33} \\
& ContrastPool  & \mstd{63.94}{4.62}  & \mstd{70.67}{6.55} & \mstd{70.78}{3.31} & \mstd{63.35}{2.68} & \mstd{51.32}{2.34} & \mstd{50.34}{2.28} \\
& IBGNN         & \mstd{64.57}{5.98}  & \mstd{77.00}{6.27} & \mstd{75.22}{5.52} & \mstd{69.43}{4.67} & \mstd{54.56}{3.40} & \mstd{54.19}{3.78} \\
& Contrasformer & \mstd{59.90}{11.46} & \bestmstd{78.67}{5.81} & \secondmstd{75.81}{4.57} & \mstd{66.89}{3.75} & \mstd{53.24}{3.68} & \mstd{51.77}{4.98} \\
\midrule
\parbox[c]{1.65cm}{\centering\itshape Ours}
& MPP-GNN & \bestmstd{77.84}{6.24} & \secondmstd{78.33}{4.56} & \bestmstd{79.48}{3.10} & \bestmstd{72.80}{3.21} & \bestmstd{60.43}{1.68} & \bestmstd{60.13}{1.02} \\
\bottomrule
\end{tabular}
\end{table*}

\subsection{Experimental Setting}

We performed five-fold cross-validation. In each split, three folds were used for training, one for validation, and one for testing, yielding a 3:1:1 ratio. At the outer level, we train the meta-optimization procedure for 50 meta-epochs, with 8 meta-weight samples drawn at each meta-epoch. The inner-level backbone GCN has hidden dimensions $(64, 32, 16)$ followed by global mean pooling and a linear classification head, and is optimized with Adam~\cite{adam} (initial learning rate $9\times 10^{-4}$ for UK Biobank and $1\times 10^{-4}$ for ADNI). We employ focal loss~\cite{focal_loss} as $\mathcal{L}_{\mathrm{cls}}$ to mitigate class imbalance, with focusing parameter $\gamma=0.6$ for UK Biobank and $\gamma=0.25$ for ADNI. All models are trained for up to 300 epochs with early stopping using a patience of 50 epochs. For affinity propagation, the maximum number of iterations is set to 3200 and the damping factor is set to 0.88. If the algorithm fails to converge, we perform a second trial with the maximum number of iterations increased to 5000 and the damping factor increased to 0.95. For both datasets, we fix $\lambda_{deg}$ and $\lambda_{sp}$ to 0.1.

\subsection{Comparison with Baselines}
We compare MPP-GNN with a diverse set of baselines, including (1) conventional machine learning methods: Support Vector Machine (SVM) and Random Forest; (2) general-purpose GNN backbones: GCN~\cite{GCN}, GAT~\cite{GAT}, GIN~\cite{GIN}, and GraphSAGE~\cite{GraphSAGE}; (3) generic graph pooling methods: DiffPool~\cite{diffpool}, TopKPool~\cite{GraphUNet}, and SAGPool~\cite{SAGPool}; and (4) brain-network-specific graph models: BrainGNN~\cite{BrainGNN}, BrainNetCNN~\cite{BrainNetCNN} and BrainNetTF~\cite{BrainNetTransformer}, ContrastPool~\cite{ContrastPool}, IBGNN~\cite{IBGNN}, and Contrasformer~\cite{Contrasformer}.

The results are summarized in Table~\ref{tab:main_results}. For both
datasets, MPP-GNN achieves the best AUC and F1, which demonstrates its
strong discrimination across decision thresholds. This is particularly
important in neuroimaging classification, where class imbalance and
threshold sensitivity can make accuracy alone less informative. For the UK
Biobank dataset, MPP-GNN achieves the best AUC and F1 score, and achieves the second-highest accuracy. Although Contrasformer has the highest accuracy, its substantially lower AUC suggests that this accuracy does not fully reflect ranking quality under class imbalance.

For the ADNI dataset, MPP-GNN outperforms all baseline models and achieves state-of-the-art results across all evaluated metrics. This result is notable because ADNI poses a more challenging three-class classification problem, in which MCI lies along a progressive continuum between CN and AD. This results in highly ambiguous class boundaries. In this setting, we observe that conventional ML methods, particularly SVM, remain competitive. This suggests that greater model complexity alone does not necessarily lead to improvement in class separation or generalization. Even deep models designed for brain networks can still suffer from overfitting when trained on limited and noisy neuroimaging data. This is particularly evident on ADNI, where several over-parameterized baselines degrade sharply. On the other hand, this also highlights the effectiveness of our methods. Our MPP-GNN combines subject-adaptive modular structure learning with learnable edge refinement, which helps the model learn neurologically meaningful topological patterns. The consistent improvement over both conventional baselines and specialized brain-network models indicates that robust neuroimaging classification may benefit from structural inductive biases that explicitly account for subject heterogeneity and noisy functional connectivity.

\subsection{Analysis}
In this section, we will analyze the reproducibility and disease-related patterns captured by PERM and AP-HPM.

\begin{figure*}[!t]
\centering
\includegraphics[width=0.80\textwidth]{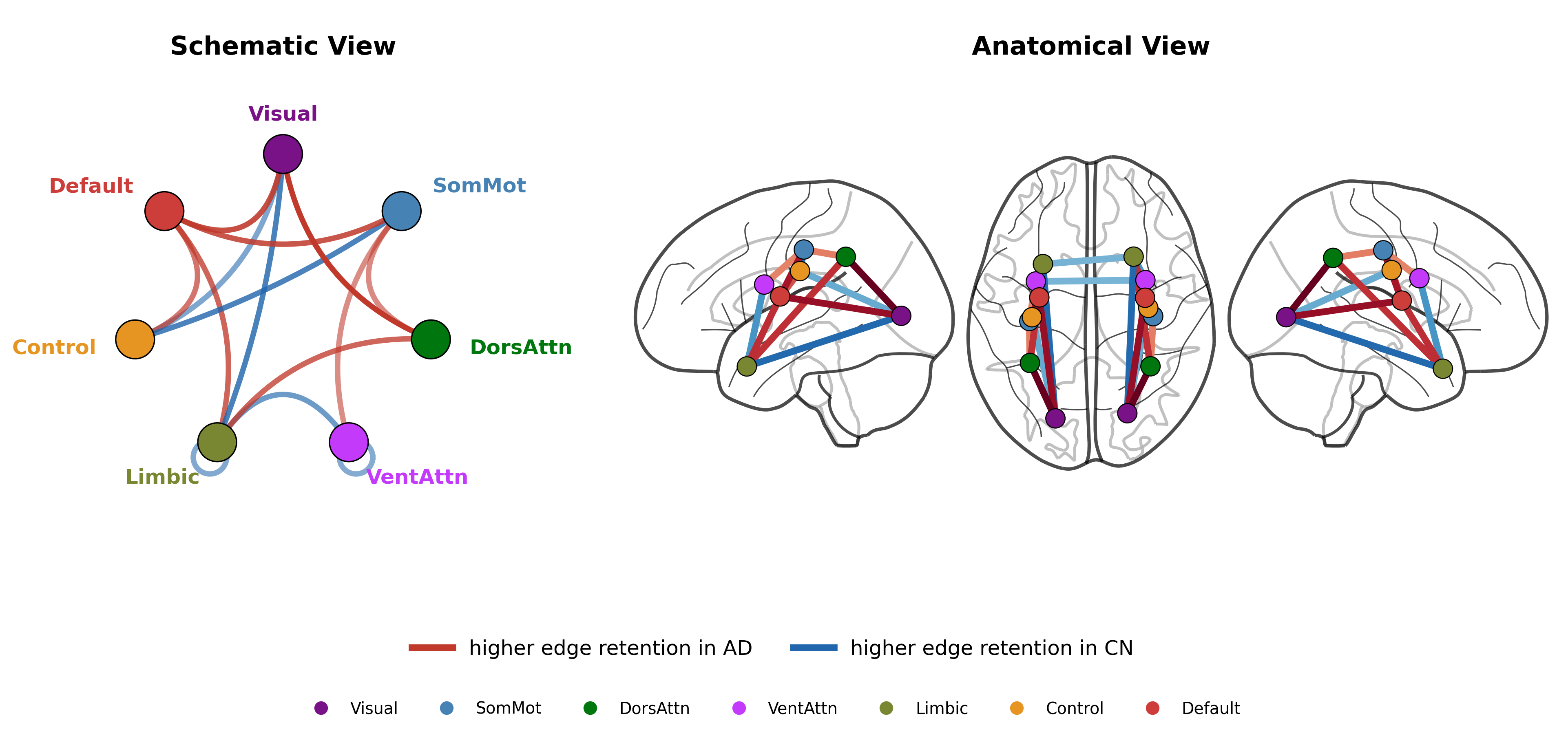}
\caption{
The chord diagram (left) visualizes the network pair that has consistently high system-level connections in most evaluated models. The glass brain (right) depicts the connections in another view with topological relations.
}
\label{fig:chord}
\end{figure*}

\subsubsection{Edge Refinement}
Edges with higher logit values, and thus higher edge-retention probabilities $P_{ij}^{(l)}$, are more likely to be preserved. We restrict the following analysis to the first-layer probability matrix $P_{ij}^{(1)}$, whose entries correspond directly to connections in the input FC matrix. We first verify that PERM learns a stable edge-keeping pattern rather than fold-specific noise. Across all 10 models evaluated (5 folds from UKB and 5 folds from ADNI), the rankings are highly consistent ($\rho=0.90$ within UKB, $\rho=0.96$ within ADNI, and $\rho=0.70$ in Spearman's rank correlation across the two datasets). This suggests that PERM learns a reproducible high-retention edge pattern under retraining, providing a stability check for the learned edge-retention readout before interpreting disease-related group-level contrasts.

An edge whose logit value is consistently higher in one group (e.g., AD) than the other (e.g., CN) may indicate a connection that is informative for disease diagnosis. Under a two-sided Mann-Whitney U test with the false discovery rate (FDR) correction, no individual functional connection shows a significantly different edge-retention probability $P^{(1)}_{ij}$ between AD and CN. However, at the Yeo-7 system level, we observe consistent group differences in edge retention between communities. Across the 10 evaluated models, the Somatomotor-Default and Visual-Default blocks show consistently higher edge-retention ranking in AD than in CN, whereas the Visual-Limbic block shows higher retention in CN than in AD (all $p<0.05$ in ADNI). These patterns are broadly consistent with previous reports of altered sensory and default-network connectivity in AD~\cite{seg-loss-zhang, vis-dmn-singh, ad-visual, default}. In Fig.~\ref{fig:chord}, we visualize the system-level connections whose rankings are consistently higher in one group than the other. Together, these effects indicate a de-differentiated coupling between sensory systems and the default network alongside reduced limbic integration.

\begin{figure*}[!t]
\centering
\subfloat[Consensus AP-HPM communities and the Yeo-7 reference.]{
\includegraphics[width=0.42\textwidth]{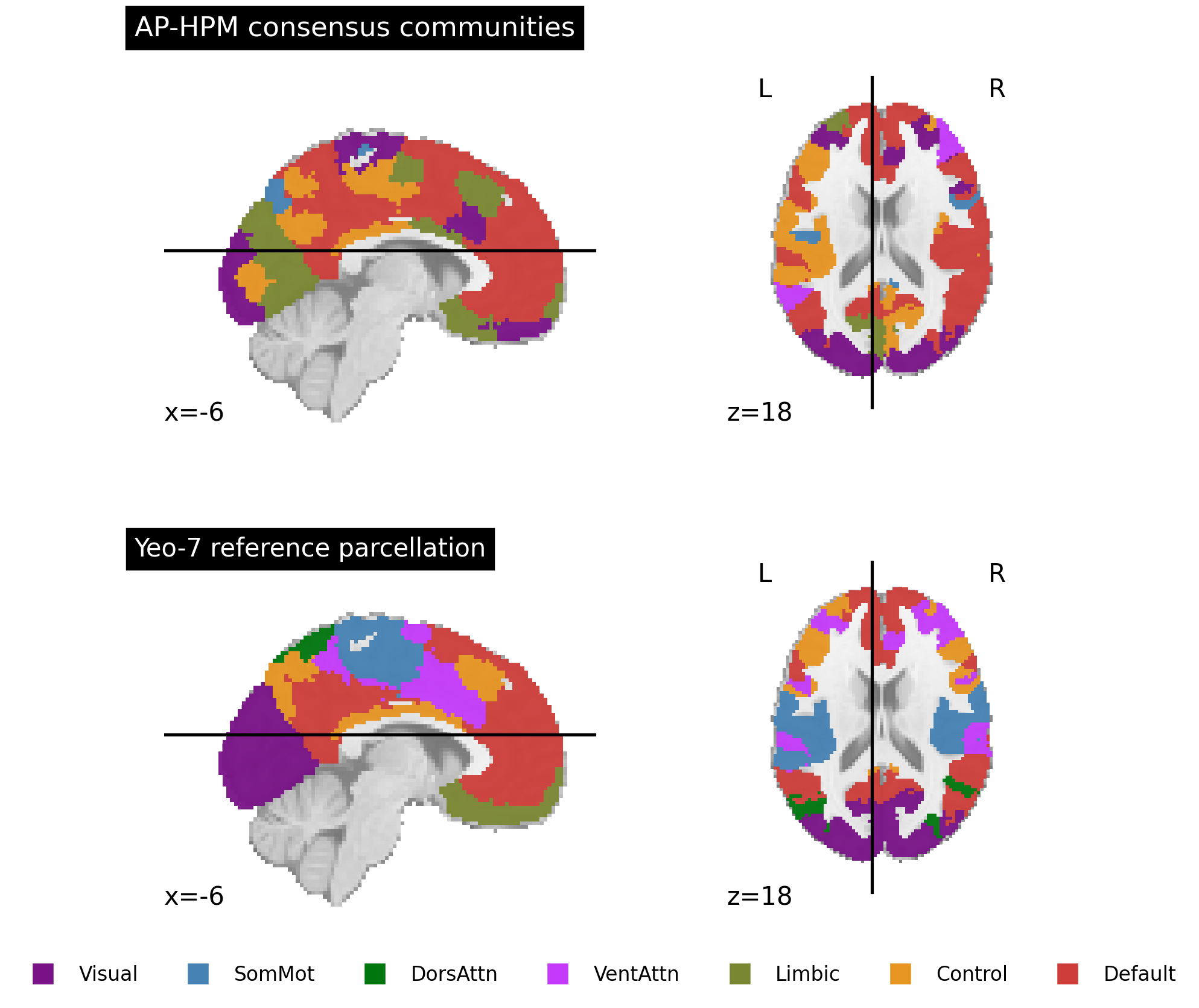}
\label{fig:aphpm_vs_yeo}}
\hfill
\subfloat[Loss of within-network co-grouping in AD relative to CN.]{
\includegraphics[width=0.42\textwidth]{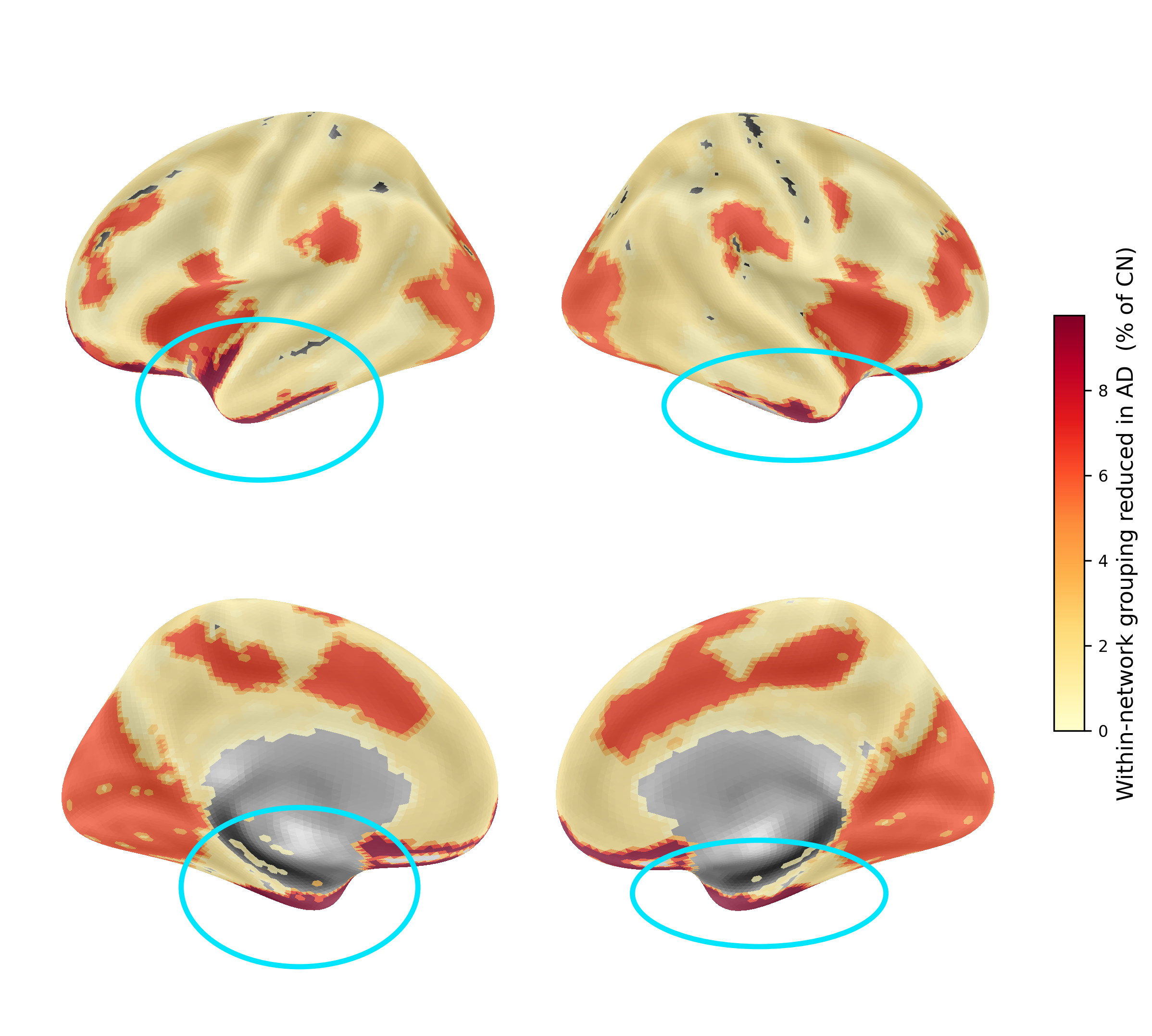}
\label{fig:dediff_simple_circle}}
\caption{AP-HPM interpretability. (a) The consensus partition is obtained from the cross-subject co-assignment matrix and recolored by majority-overlap Yeo-7 network. (b) Warmer colors indicate greater AD-related loss of within-network co-grouping; the cyan circle marks the temporal pole.}
\label{fig:aphpm_interpretability}
\end{figure*}

\subsubsection{Hierarchical Community}

To verify whether AP-HPM learns meaningful brain networks, we compute the normalized mutual information (NMI) between the first-layer partition and the Yeo-17 atlas~\cite{yeo2011}. Mann-Whitney U tests show that the learned AP-HPM clusters have a significantly higher NMI than a permutation null obtained by randomly shuffling the Yeo-17 network assignments across ROIs while preserving the network-size distribution ($p<0.001$). This indicates AP-HPM has captured some meaningful system-level structure in the brain. We further validate this by computing a co-assignment matrix, which counts the frequency with which a pair of ROIs is pooled into the same community among all subjects. Averaging the co-assignment over all subjects and clustering it agglomeratively into 17 clusters produces a consensus partition whose NMI against Yeo-17 exceeds that of $99.7\%$ of the individual subjects. This suggests that AP-HPM can capture a stable pattern of the brain network instead of subject-specific random partitions. For clear visualization, we recolor the partitions by each community’s majority Yeo-7 network in Fig.~\ref{fig:aphpm_vs_yeo}.

As with the individual edges in PERM, the subject-level AP-HPM partition pattern does not separate AD and CN cohorts by itself, indicating that the edge refinement and subsequent representation learning are necessary for disease diagnosis. However, at the system level, the partitions of AD subjects align significantly less with the canonical networks (lower NMI against Yeo-17 than those of CN). The fact that the direction (AD $<$ CN) is consistent across all 10 evaluated models and significant in ADNI ($p \le 0.006$) indicates dedifferentiation in AD. Additionally, Fig.~\ref{fig:dediff_simple_circle} visualizes the difference in within-network co-grouping between AD and CN, i.e., how frequently ROIs belonging to the same Yeo-7 network are pooled together. The circled region highlights the temporal pole, which is a limbic-associated cortical region reported to be affected early in AD~\cite{early-ad-berrone, early-ad-braak}.

\subsection{Ablation Studies}

In this subsection, we validate the effectiveness of the key components in MPP-GNN, including the AP-HPM module, the PERM module, the bilevel optimization framework, and the loss function. All ablation experiments are conducted on the UK Biobank dataset using 5-fold cross validation.

\subsubsection{MPP-GNN Modules} 

We progressively add the proposed modules onto a plain GCN backbone to isolate the contribution of each component. Results are reported in Table~\ref{tab:ablation_module}. Disabling all three components (AP-HPM, PERM, and bilevel optimization) reduces the model to a three-layer GCN. The performance is close to the GCN baseline in AUC reported in Table~\ref{tab:main_results}. Enabling AP-HPM alone yields a moderate improvement, indicating that community-guided hierarchical pooling provides additional structural information beyond flat message passing. We then further incorporate PERM into the framework. The performance improves substantially, which highlights the severity of noise in the original FC matrices and the effectiveness of probabilistic edge refinement in suppressing noisy connections. Finally, enabling the bilevel optimization framework brings a further improvement to $77.84\%$ AUC and $78.33\%$ accuracy. This demonstrates that the outer-level meta-optimization can discover superior community partition strategies that in turn benefit edge denoising and representation learning through the coupled pipeline.

\subsubsection{Loss Function} The loss function $\mathcal{L}_{\mathrm{edge}}$ encourages the edge-retention probabilities $P_{ij}^{(l)}$ to move away from $0.5$, thus sharpening the distinction between retained and suppressed edges. We test the effectiveness of the design of the loss function on both a complete MPP-GNN framework and an ablated version without the outer-level optimization. As shown in Table~\ref{tab:ablation_loss}, $\mathcal{L}_{\mathrm{edge}}$ plays an important role in boosting the model performance in both frameworks.

\begin{table}[t]
\caption{Ablation Study on Modules on UK Biobank dataset. The best result is highlighted in \textbf{bold}.}
\label{tab:ablation_module}
\centering
\footnotesize
\setlength{\tabcolsep}{6pt}
\setlength{\aboverulesep}{0.15ex}
\setlength{\belowrulesep}{0.15ex}
\renewcommand{\arraystretch}{0.94}
\begin{tabular}{
    >{\centering\arraybackslash}p{1.2cm}
    >{\centering\arraybackslash}p{1.0cm}
    >{\centering\arraybackslash}p{0.6cm}
    >{\centering\arraybackslash}p{1.8cm}
    >{\centering\arraybackslash}p{1.8cm}
}
\toprule
AP-HPM & PERM & BO & AUC & ACC \\
\midrule
            &              &            & \mstd{54.07}{3.50} & \mstd{66.33}{7.41} \\
\checkmark  &              &            & \mstd{60.38}{6.55} & \mstd{68.00}{13.06} \\
\checkmark  & \checkmark   &            & \mstd{70.00}{7.28} & \mstd{74.00}{2.26} \\
\checkmark  & \checkmark   & \checkmark & \bestmstd{77.84}{6.24} & \bestmstd{78.33}{4.56} \\
\bottomrule
\end{tabular}
\end{table}

\begin{table}[t]
\caption{Ablation Study on Loss function on the UK Biobank dataset. The effect is evaluated with and without bilevel optimization (BO). The best result is highlighted in \textbf{bold}.}
\label{tab:ablation_loss}
\centering
\footnotesize
\setlength{\tabcolsep}{6pt}
\setlength{\aboverulesep}{0.15ex}
\setlength{\belowrulesep}{0.15ex}
\renewcommand{\arraystretch}{0.94}
\begin{tabular}{
    >{\centering\arraybackslash}p{0.9cm}
    >{\centering\arraybackslash}p{1.1cm}
    >{\centering\arraybackslash}p{0.6cm}
    >{\centering\arraybackslash}p{1.8cm}
    >{\centering\arraybackslash}p{1.8cm}
}
\toprule
$\mathcal{L}_{\mathrm{cls}}$ & $\mathcal{L}_{\mathrm{edge}}$ & BO & AUC & ACC \\
\midrule
\checkmark &              &             & \mstd{62.93}{5.42} & \mstd{73.33}{4.08} \\
\checkmark & \checkmark   &             & \mstd{70.00}{7.28} & \mstd{74.00}{2.26} \\
\checkmark & \checkmark   & \checkmark  & \bestmstd{77.84}{6.24} & \bestmstd{78.33}{4.56} \\
\bottomrule
\end{tabular}
\end{table}

\section{Conclusion}

In this paper, we proposed MPP-GNN, a GNN-based framework for brain network classification with a focus on Alzheimer's disease. It adopts a bilevel optimization framework to jointly optimize the graph partition strategy and graph representations, and refines noisy connectivity patterns by using the discovered community structure as a subject-specific prior to guide inner-level graph structure learning. Because both AP-HPM and PERM are interpretable by construction, MPP-GNN produces explanations directly from its learned parameters rather than from post-hoc attribution. Experimental results on two public datasets show that MPP-GNN consistently outperforms the compared baselines and achieves strong classification performance. In addition, we demonstrate that the proposed AP-HPM discovers community partitions that align with the canonical Yeo networks, while PERM learns reproducible edge-retention patterns across folds and datasets. Both modules further converge on a consistent network-level dedifferentiation in AD, in agreement with the established literature. In future work, we will investigate the integration of non-imaging information, test on additional brain disorder datasets, improve the computational efficiency of the framework, and further explore other graph partition strategies and model designs.

\bibliographystyle{IEEEtran}
\bibliography{references}

\end{document}